# Automated Weed Detection in Aerial Imagery with Context

Delia Bullock[1]*, Andrew Mangeni[1]*, Tyr Wiesner-Hanks[2], Chad DeChant[1], Ethan L. Stewart[3], Nicholas Kaczmar[4], Judith M. Kolkman[3], Rebecca J. Nelson[2,3], Michael A. Gore[2], and Hod Lipson[4]

## Abstract

In this paper, we demonstrate the ability to discriminate between cultivated maize plant and grass or grass-like weed image segments using the context surrounding the image segments. While convolutional neural networks have brought state of the art accuracies within object detection, errors arise when objects in different classes share similar features. This scenario often occurs when objects in images are viewed at too small of a scale to discern distinct differences in features, causing images to be incorrectly classified or localized. To solve this problem, we will explore using context when classifying image segments. This technique involves feeding a convolutional neural network a central square image along with a border of its direct surroundings at train and test times. This means that although images are labelled at a smaller scale to preserve accurate localization, the network classifies the images and learns features that include the wider context. We demonstrate the benefits of this context technique in the object detection task through a case study of grass (foxtail) and grass-like (yellow nutsedge) weed detection in maize fields. In this standard situation, adding context alone nearly halved the error of the neural network from 7.1% to 4.3%. After only one epoch with context, the network also achieved a higher accuracy than the network without context did after 50 epochs. The benefits of using the context technique are likely to particularly evident in agricultural contexts in which parts (such as leaves) of several plants may appear similar when not taking into account the context in which those parts appear.

## 1 Introduction

The accurate identification of weeds is critically important to ensure proper weed control practices can be implemented in maize cropping systems. When combined with the tools of digital agriculture to more cost-effectively and efficiently control grass or grass-like weeds (hereafter weeds), image-based weed detection offers the potential to support the precise application of herbicides based on both the weed species and its location in the field. However, it remains a challenge to unambiguously identify specific weed species with field imagery collected from maize fields.

Object detection is a task within computer vision that involves both finding and classifying objects from a set of object classes in an image. Therefore, object detection can be seen as a two-part problem that can be broken down into classification and localization computer vision tasks.

Image classification is one of the most familiar and popular tasks within computer vision. Feature detection and pattern recognition are at the core of image classification with convolutional neural networks, which are essential to many deep learning computer vision tasks. The innovation within these networks came from the fact that they could learn both higher level and lower level patterns of objects without the need for manual feature engineering. These learned features and patterns could then be used to differentiate between object classes.

[1]Department of Computer Science, Columbia University, New York, NY 10027, USA
[2]Plant Breeding and Genetics Section, School of Integrative Plant Science, Cornell University, Ithaca, NY 14853, USA
[3]Plant Pathology and Plant-Microbe Biology Section, School of Integrative Plant Science, Cornell University, Ithaca, NY 14853, USA
[4] Department of Mechanical Engineering and Institute of Data Science, Columbia University, New York, NY 10027, USA

Localization is a computer vision task that involves finding the location of an object within an image. After the object is found, there are a number of different options for marking the boundary of its location, but one of the most widely used choices is a bounding box.

The difficulty of the object detection task stems from the fact that it is a combination of the image classification and localization tasks. Therefore, any network used for this task must be able to both locate objects of interest and classify them within the image. This is challenging because both of these tasks increase each other's difficulty. For example, the images found within widely used image classification datasets often contain well centered and easy to identify objects. In addition, localization of objects requires that they are properly classified with the correct object class. A few of the other difficulties in this task are that objects from the same class can vary in size and shape within the images and there can be a variable number of instances of the objects.

Though many of the difficulties within object detection have been solved with approaches that have achieved state-of-the-art results, a problem that still persists in object detection is the problem of scale. Because convolutional neural networks learn from feature detection, these networks often run into problems when objects in different classes have similar features. One common scenario when this arises within the object detection task is when the scale of the image regions being fed to the neural network prevents distinct features from being seen in objects of interest. At a smaller scale, similarities in patterns can create edge cases where different objects show similar characteristics. These cases cause errors in classification and lower the accuracy of the network. We will call this problem the small-scale similarity problem. Since images within the object detection task are taken at varying distances and objects within these images can vary in size and shape, there are many places within an image where the small scale similarity problem can present difficulties.

Within the weed detection case study, we observed many scenarios where the small-scale similarity problem occurs. As shown in Figure 1, it is hard to differentiate what is a weed (i.e., an undesirable plant that is not maize) from what is not. This is due to the fact that leaves of certain weeds and maize can all look the same at small scales. Even humans, who are incredible at pattern recognition, can get confused when a small-scale gives a limited view of an object. Therefore, it is quite understandable that neural networks would also have similar trouble. Also, this shows that it is not a problem that can be solved by simply changing network architecture. Another seemingly intuitive approach would be to feed neural networks larger image regions so that small-scale is not a problem. However, this creates added difficulties in learning features of smaller objects and accurately locating them.

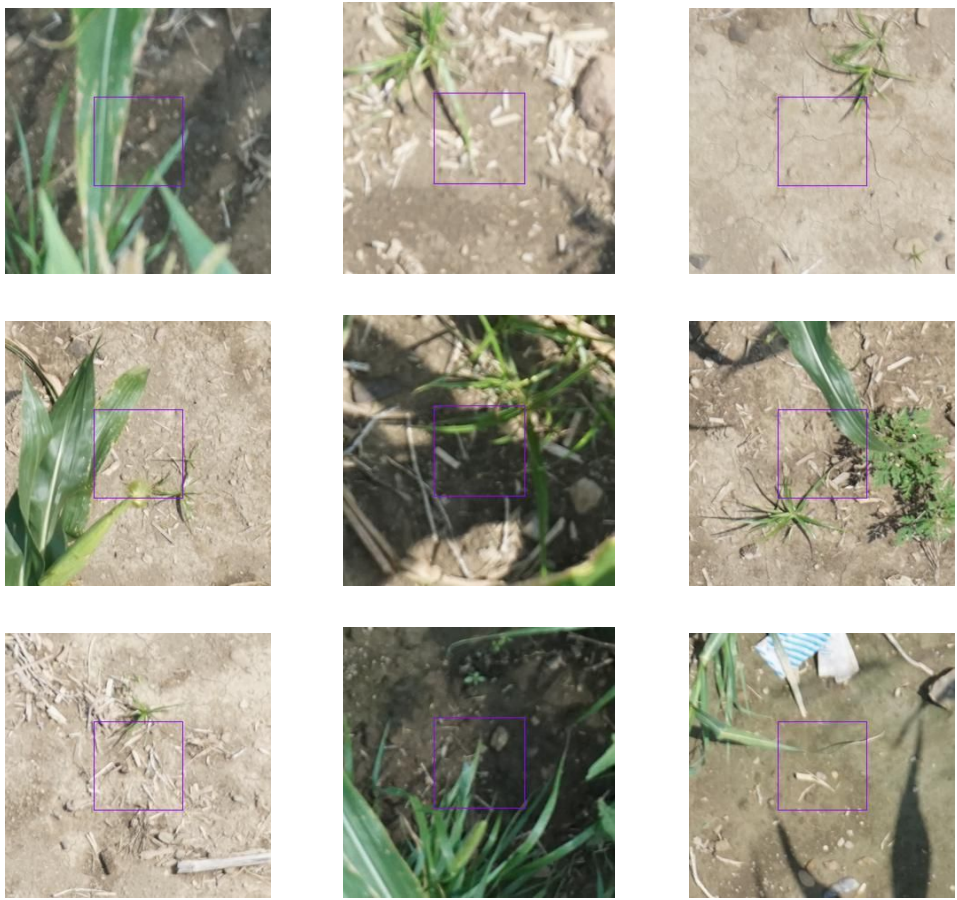

Figure 1: Example of images that were not identified without context, but were identified by the network when context was added. The added context is the area outside of the purple square.

We address this problem using a little-used technique involving adding "context," which allows our network to view images that are labeled on a smaller scale but are accompanied by surrounding background area when ingested by the network.

Using the context technique, we are able to significantly improve the performance of a convolutional neural network-based system to detect the presence of weeds in field imagery of maize plants.

## 2  Related work

Significant useful work in data augmentation and preparation for the task of object detection can be found in [2]. Ever since convolutional neural networks began being applied to unconstrained environments, the issue of feeding the network well-centered or meaningful images has presented difficulty. To solve this, [2] combined multi-scale with pre-existing multi-crop techniques to scan input images at varying locations and resolutions. Not only is the network able to see the target from multiple frames, but the addition of multi-scale provides the network with valuable background image information. This technique greatly improved their networks' accuracy; however, it requires a complex architecture to digest the images at different resolutions. Furthermore, given that it is an ensemble method, it is computationally and time expensive. In [3], researchers found that in practice, these combined techniques are often too complex and computationally demanding to be worthwhile.

Different forms of context have previously been explored within the object categorization, localization, and detection tasks. Our approach is very similar to that used in [12], which we found after independently developing our technique. In [10], researchers used semantic context, which involves understanding the overall idea of what is happening in an image by looking at the relationships between objects in it, to reduce ambiguity when classifying each individual object within the image and aid with the object categorization task. A similar idea is used in [11] and applied to the object recognition task. Here, the central idea is that knowing what objects are around a target object can help with understanding its context, which improves detection. Context can also be used to improve object recognition by determining the setting of an image.

By focusing not only on what an object is but where it is found, researchers in [9] were able to use familiar locations to categorize new locations and then use that information as a prior within object detection. Therefore, knowing that an image is set in a kitchen allows a network to more likely classify an object as a lemon rather than a tennis ball. Context has been especially useful within weakly supervised and even unsupervised tasks. Detailed image annotation is difficult and time intensive, making it almost impossible for some tasks to have this kind of supervision. However, researchers have found that convolutional neural networks can still provide accurate classification and approximate localizations of objects while relying only on image-level labels [7]. Other researchers found that adding context in instances of weakly supervised localization, where only image level supervision was used, resulted in improvements from the prior benchmarks [6]. Here context was defined as everything else in the border that surrounds a target picture. In [12], context was used as a supervisory signal for an unsupervised learning task, which allowed the task to be turned into a weakly supervised task.

In real-world applications, finding the middle ground between giving the network too much or too little information is very difficult. More recently, researchers have approached this issue with a mechanism known as attention. In theory, attention allows the network to focus on what it considers important by first viewing the image in a series of multi-resolution crops or "glimpses." Each glimpse is fed into a deep recurrent neural network to determine the next glimpse and finally classify the image. These networks are especially helpful in fine-grained classification [1], which is highly related to our application since leaves of various weeds and maize can look so similar they are almost indistinguishable especially when obstructed. Although they have achieved state of the art results within object detection [8], recursive neural networks are notoriously difficult to train and attention models in general may be hard to implement.

In some real world applications, it may be both more feasible and beneficial to use context. Context is actually used in industry inspection, where defects in texture must be detected in real time. The context used here takes a larger crop of a target image and then appends it next to the target image after the dimensions are fit [5]. Here it is easy to see the benefit that context has, because it can help detect minute differences in texture that would possibly be overlooked if context was not taken into account. Also, since context does not require several different crops of the same section it can provide detections quickly. In [4], we see an instance of object detection aiding with disease detection within maize plants. This is a scenario that is very similar to the case study on weed detection in maize fields in which we tested the application of the context technique. Therefore, this could also be an instance where the addition of context may lead to improved accuracy.

## 3 Experiment

### 3.1 Data Preparation

Within the weed case study, we began with 224 aerial photos of maize field experiments. Images were taken using a Sony A6000 mounted on a drone over four dates in August 2016. Original images were 4000x6000 pixels, taken from variable altitudes and thus showed a ground area between roughly 2x3 meters and 14x21 meters. Larger images were cropped into smaller subimages that showed roughly a 2x3 meter area on the ground. The maize experiments consisted of diverse germplasm with variable germination and vigor, such that some images showed mostly a closed maize plant canopy and others mostly bare ground and weeds. Heavy rains throughout the summer of 2016 contributed to particularly bad weed infestations.

The first step was to randomly split them into training, validation, and test sets (70%/15%/15% - 158/33/33). These large photos were then split into smaller 300x300 pixel square images by placing a grid with 300x300 pixel squares over the large photo (Figure 1), and cropping the squares based on the gridlines. It was important to first split the large photos into training, validation, and test sets before cropping them into smaller square photos. This ensured that there were no 300x300 square images cropped from the same large picture within multiple sets, which would incorrectly bias our results. After splitting all the full-size images into 300x300 squares, we manually labeled each of these 300x300 square images as weed or non-weed.

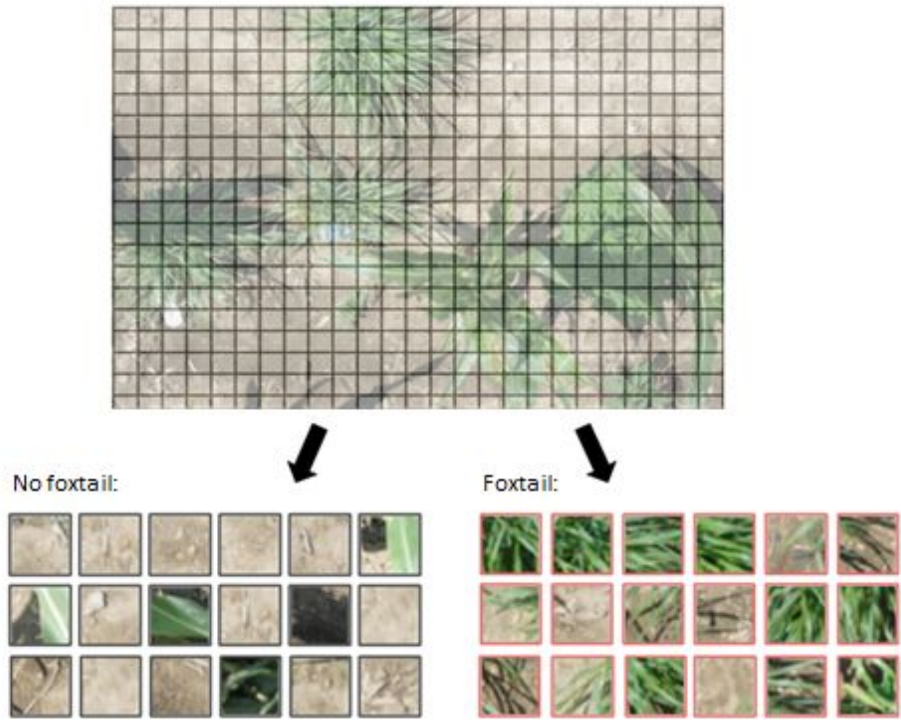

Figure 2: Example of using a grid to crop a full-size image into smaller 300x300 pixel squares.

In our study, full size picture, full size image, large picture, or large image refers to the 224 large aerial drone photos of the maize fields. Additionally, 300x300 square, square, smaller picture, or smaller image refers to the smaller 300x300 pixel photos that were cropped from the large photos by overlaying a grid.

Since even the slightest data impurities would affect both the features our network learned and the accuracies of the network, we manually labelled each 300x300 square.

After manual labelling, it was evident that there were eight times more photos containing non-weed than weed instances. In order to balance this disproportion, we created and employed a data jittering technique to augment the existing weed photos that utilized our grid structure. This technique was similar to data augmentation techniques used within computer vision to increase image data. We jittered the weed photos until the data were balanced. We took these new data into account (jittered and non-jittered) and re-split the large photos into training, validation, and test sets. This made it much less likely for the validation set to have outliers in relation to the training set, making validation error more reliable.

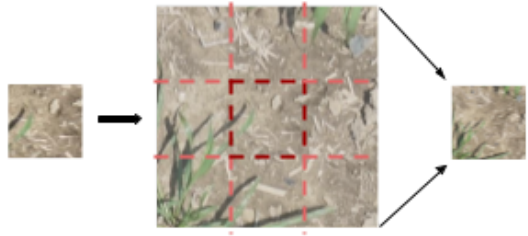

Figure 3: Illustration of photos before and after context-adding. The picture on the far left is the input, labelled as having a weed. The middle picture shows the input image with a 300 pixel width border added around it. This new larger photo keeps the input image's original label, which in this example contained a weed (foxtail), then its dimensions are resized back down to 300x300 pixels. The picture on the far right is the result of adding context and resizing.

The technique we created and applied to our data is called context. This technique involves first taking a 300x300 pixel square and adding a 300 pixel-width border of its surrounding to the central square. The new 900x900 pixel square image is then labeled with the original 300x300 central square's label. Using the network on the original central square's label rather than relabelling the larger square is important and one of the things that makes this technique different from just cropping larger squares. Labelling this way allows us to keep localization at a smaller scale, which is essential to object detection, while providing our network an opportunity to learn more features. The newly labelled image is 900x900 pixels, so we resize this new square back down to 300x300 pixels. The network is now trained on these resized images instead of the original images. Figure 3 shows our technique of adding context around a central square.

To test the effect of adding context to the data, we created three datasets: a no context dataset in which this technique is not applied, a full-stretched context dataset, and an edge-stretched context dataset. The last two datasets will be further explained when we detail their creation. These three datasets were created with identical training/validation/test/ splits and identical random jittering.

To create the full-stretched context dataset, we first added context to the images. This dataset is called full-stretched context because of the way we chose to handle the situation when a central square was located on the edge of the full-size image. Whenever there is not a complete 300 pixel-width border on one or more sides of the central square, we take whatever number of pixels are left from this specific side of the central square to the edge of the full-size image instead. Because the number of pixels we add on this side is not exactly 300, we end up having a rectangle formed rather than a 900x900 pixel square image. Therefore, when we reduce the dimensions of the new context rectangle down to a 300x300 square we get a square that has an image that is slightly stretched because it has been shrunk unevenly. Although it is possible for the weed within a central square to be moved outside the center of the reduced image square, we allowed this stretching to happen rather than discarding any squares on the edges because we hypothesized that this would give us different data that would make our network more robust in a way similar to other warping techniques people often use to augment their data.

The next dataset we created was the edge-stretched context dataset, in which we added context to the images in a similar way to the full-stretched context dataset. However, unlike in the full-stretched context dataset, whenever there is not a complete 300 pixel-width border on one or more sides of the central square, the pixels left on the border of that side would be stretched to 300 pixels. Therefore, the larger image created when context is added is still a square. This method of "edge-stretching" allowed us to add a 300 pixel width border around the original central square and shrink the dimensions down to 300x300 pixels without distorting the central square (which was the case in the full-stretched dataset). Therefore, any weed that was previously in the center square was still in the center square after the context was added and the dimensions were reduced. The same was true for non-weed instances. This edge-stretched method allowed us to add context to the data without full-stretching the final 300x300 square that was passed to the network. Therefore, we could verify whether adding context or the stretching that occured in the full-stretched case improved the network's results. At the end of the data preparation we had three equal number datasets that differed only in application of "context."

### 3.2 Network Design

The focus of this experiment was to observe the effects of adding context, so we designed a very basic network architecture that would not be biased towards one type of data. This network is illustrated in Figure 4. It contains six convolutional layers: two that extract 32 filters, then two that extract 64 filters, and finally two that extract 128 filters. After each set of two layers there is a 2x2 max pooling layer. The number of filters extracted increases as the data are shrunk through max-pooling, as is standard practice. The convolutional layers are followed by a fully connected layer of 64 nodes. The network ends in a single binary node that outputs 1 for weed and 0 for non-weed, the same labels we used while tagging the data. The hyperparameters learning rate and decay were then slightly adjusted using full-stretched validation data. This is a very standard network that could most likely be optimized much more, but is kept simple to provide meaningful results.

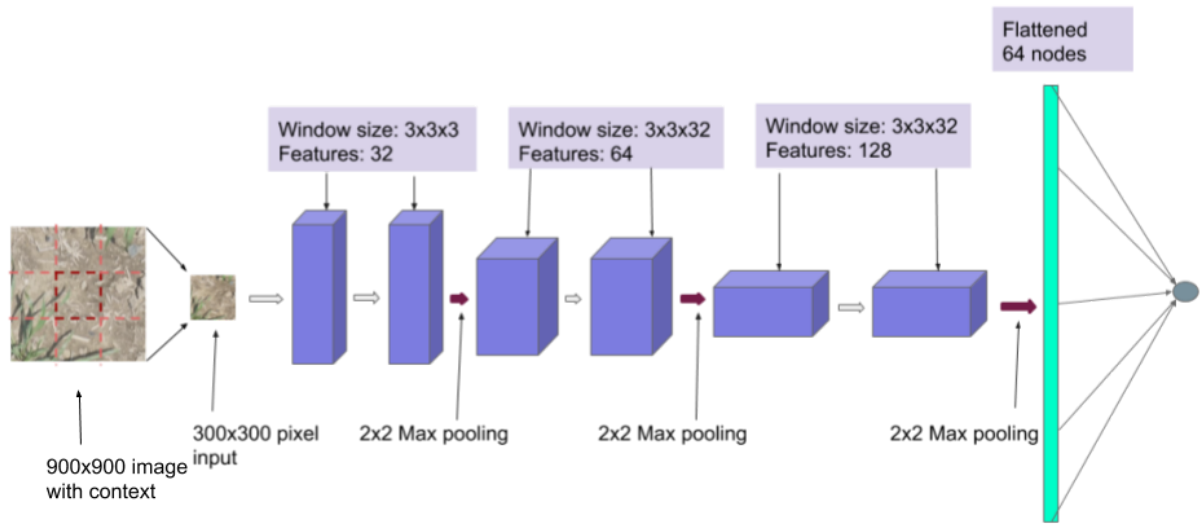

Figure 4: Diagram of the seven-layer deep network used in this experiment.

## 4 Results

The model was trained on each of the three datasets described in section 2.1. For each dataset, we ran three independent runs. Each run was 50 epochs and took approximately 20 hours. Each set of three runs were averaged together and their validation accuracies are compared in Figure 5. The errors from the epoch with the highest validation accuracy for each respective data preparation method are also compared in Figure 6. The highest validation accuracies were 94.6%, 96.3%, and 97.1% for no context, full-stretched, and edge-stretched respectively. Adding context cut the error in half without any additional training time. From Figure 5, it seems that for both context and non-context methods, learning was saturated around epoch 25, but context methods were able to significantly outperform data without context. From Figures 5 and 6, it appears that of the two contextualizing methods, edge-stretching was consistently more effective than full-stretch. We will refer to the highest performing model as the Edge-Stretched Context (ES-Context) Model.

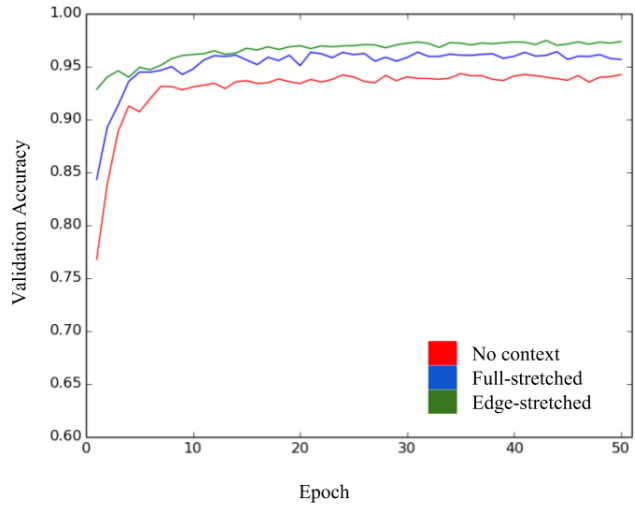

Figure 5: Accuracies of three different models over 50 epochs. Each line is the average of three runs.

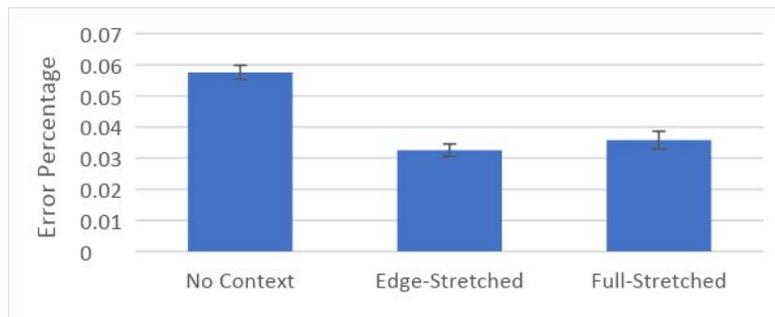

Figure 6: Errors of three different models (averaged over three runs) trained on no context, edge-stretched, and full-stretched data.

The ES-Context Model was then tested on the held-out test data set. The results for this test compared against the no-context model are shown in Table 1. This was the only time the held-out dataset was tested.

Table 1: Performances on held-out test dataset

|  | No context | ES-Context |
| --- | --- | --- |
| Accuracy | 92.9% | 95.7% |
| Precision | 61.9% | 75.5% |
| Recall | 88.5% | 88.6% |

The trained models can be used to create heat maps of the original drone images by sliding the input window across the full image and implementing the appropriate stretching or resizing dynamically when needed. Figure 7 shows one such heat map.

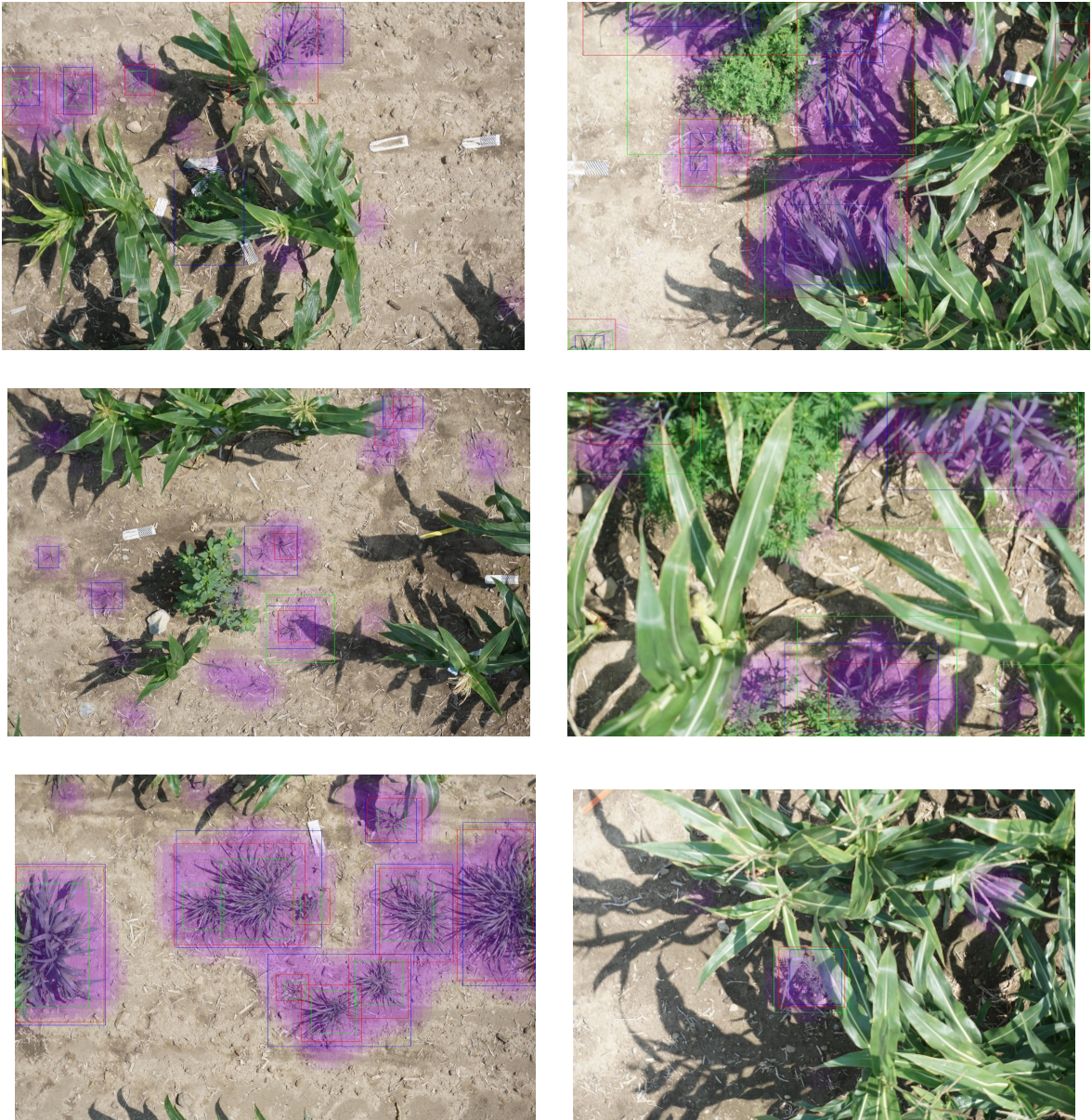

Figure 7: Heat map indicating detected foxtail and yellow nutsedge weeds generated by ES-Context Model. Mechanical Turkers' responses are shown as colored squares and overlaid on the heat map. All of the instances of purple on the heat map are correct identifications except the one in the bottom right and the one below the tassel (male inflorescence of the maize plant).

A human baseline comparison of the ES-Context Model is shown in Figure 8. The human baseline was created using Mechanical Turk. Full sized drone images were shown to Turkers who were asked to draw boxes around all of the weeds in each photo. The Turkers' responses were compared against heat maps output by the network. The network was able to find 150% more weed patches than the human average. The network misclassified non-weed vegetation patches at a rate slightly higher than the human average. These results indicate that this network can far outperform the human average and it is much more likely to over predict weed presence than to miss weeds.

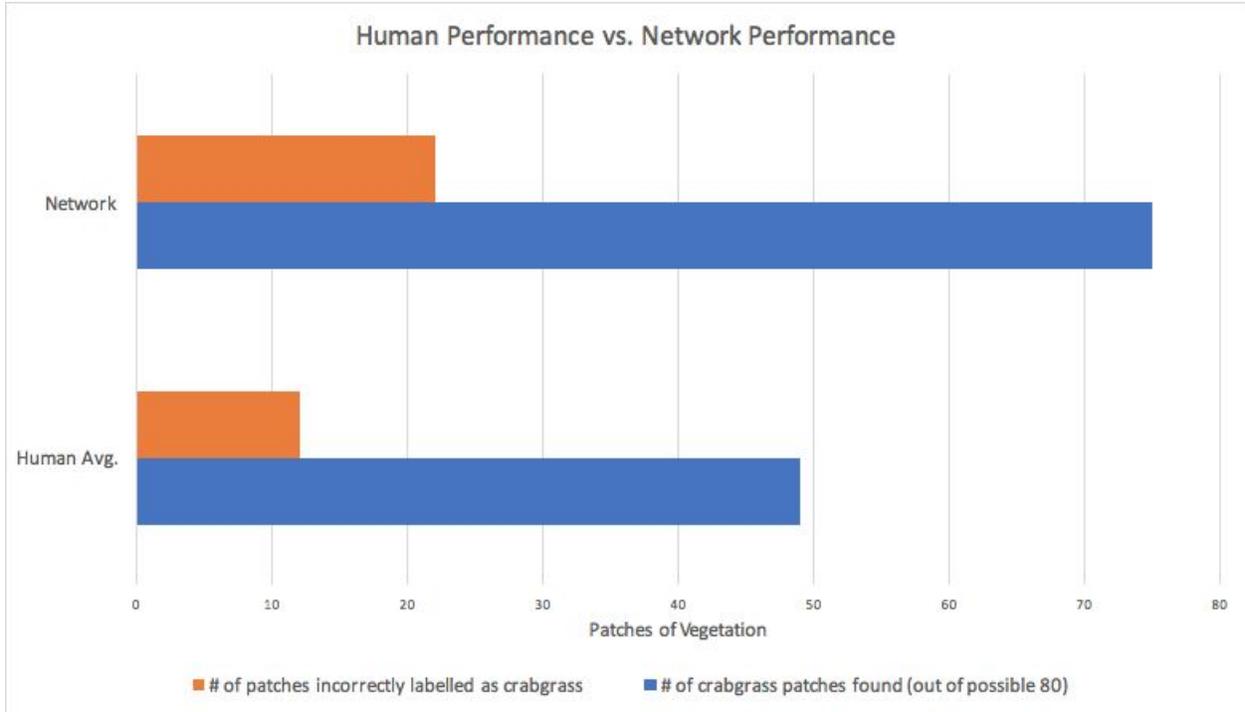

Figure 8: Graph of human average performance compared to ES-Context Model (network) performance. 14 full-sized images were evaluated that contained a total of 80 weed instances.

## 5   Discussion

In this paper, we discussed the obscure technique of enhancing data with context. This simple data augmentation method has been shown to significantly improve network performance in our weed detection case study. In Figure 9, there are photos that were misclassified by the no context network, but correctly classified once context was added.

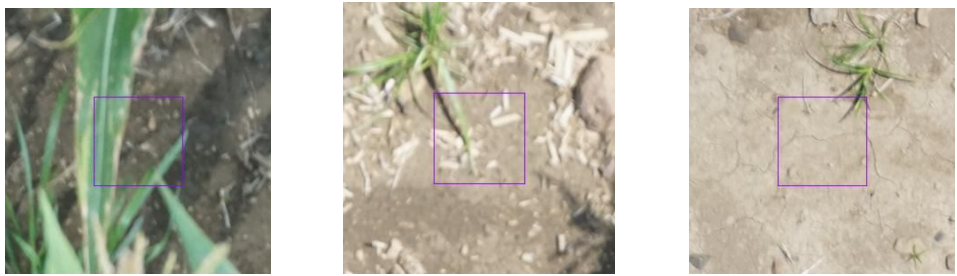

Figure 9: Examples of photos that required context to be correctly classified.

The weed case study lends itself very well to this technique since a grass or grass-like weed, the target, can be almost indistinguishable from non-targets (maize plants) in close cropped photos. Just as a human would need surrounding area to distinguish targets from non-targets, so does a neural network. The weeds also lend themselves very well to edge cases where context needs to be warped and resized. This is because vegetation often exists in many different stand sizes, orientations, stem lengths, and growth patterns. This technique might not be as fruitful for applications such as identifying cats versus dogs, where features are more clearly differentiable. However, in any applications where targets and non-targets look very similar and may be obstructed or cropped in the input data, this technique could be very impactful. Applications such as identifying medical lesions, analyzing environmental data, other vegetation classification tasks, biology applications, and many other image classification tasks could benefit a lot from context enhancement.

This technique requires photos to be resized (and in the case of edge-stretching sometimes resized more than once) slightly increasing the time needed for data preparation and classification. However, it is substantially faster and easier to implement than the popular multi-crop/multi-scale method, which approaches the same issue of trying to present the network with a properly framed target. Unlike multi-crop/multi-scale, adding context does not require the network to evaluate the same image multiple times. Thus, it is solving a similar issue in a more efficient way. It is also easier to implement and more intuitive while still reaping improved performance.

Context enhancement is also suitable for situations in which system administrators either do not feel comfortable, are not capable, or simply do not want to highly optimize and tune a network to achieve accurate results. In our experiment, the network was purposefully very standard and not optimized for one particular type of data. We believe that the network could be further optimized to realize greater improvement. However, in situations where administrators do not want to allocate resources for fine-tuning, they can easily achieve increases in accuracy by simply adding context when training.

The disproportionate weed data are also an illustrative example of when data augmentation is necessary but built-in techniques cannot and should not be used since those techniques would distort the images too much and possibly skew the data. In this case study, we designed an effective grid-based jittering method that can be used when built-in methods are unavailable.

In this case study, the addition of context appears to be a useful technique that deserves further attention beyond weed detection in field imagery. Additionally, we expanded the data input by 200% to further augment the dataset, but further research efforts are needed to determine whether it is the optimal number and if it would vary depending on the application. Overall, the techniques demonstrated in this paper have the potential to be deployed for sustainable weed control in fields of maize and other crop species.

## Acknowledgements

This work was supported by the U.S. National Science Foundation IIS-1527232 (M. A. Gore, R. J. Nelson, and H. Lipson). We thank Columbia University for providing GPU resources for this research.

## References

[1] Sermanet, Pierre, Andrea Frome, and Esteban Real. "Attention for fine-grained categorization." *arXiv preprint arXiv:1412.7054,* (2014).

[2] Sermanet, Pierre, et al. "Overfeat: Integrated recognition, localization and detection using convolutional networks." *arXiv preprint arXiv:1312.6229,* (2013).

[3] Simonyan, Karen, and Andrew Zisserman. "Very deep convolutional networks for large-scale image recognition." *arXiv preprint arXiv:1409.1556,* (2014).

[4] Dechant, Chad et al., ''Automated identification of northern leaf blight infected maize plants from field imagery using deep learning.'' Phytopathology, vol. 107, no. 11, pp. 1426–1432, (2017)

[5] Weimer, Daniel, Ariandy Yoga Benggolo, and Michael Freitag. "Context-aware Deep Convolutional Neural Networks for Industrial Inspection." 10.13140/RG.2.1.2428.3765, (2015).

[6] Kantorov, Vadim, Maxime Oquab, Minsu Cho, and Ivan Laptev. "Contextlocnet: Context-aware deep network models for weakly supervised localization." In ECCV, (2016).

[7] Oquab, Maxime, Leon Bottou, Ivan Laptev, and Josef Sivic. "Is object localization for free?–weakly-supervised learning with convolutional neural networks." In CVPR, pages 685–694, (2015).

[8] Girshick, Ross, Jeff Donahue, Trevor Darrell, and Jitendra Malik. "Region-Based convolutional networks for accurate object detection and segmentation." In TPAMI, (2015).

[9] Torralba, Antonio, Kevin Murphy, William Freeman, Mark Rubin, "Context-based vision system for place and object recognition." In ICCV, pp. 273–280. In IEEE, (2003).

[10] Rabinovich, Andrew, Andrea Vedaldi, Carolina Galleguillos, Eric Wiewiora, Serge Belongie. "Objects in context." In ICCV, pp. 1–8. In IEEE, (2007).

[11] Oliva, Aude, Antonio Torralba, "The role of context in object recognition." Trends in Cogn. Sci. 11(12), 520–527, (2007).

[12] Doersch, Carl, Abhinav Gupta, Alexei A. Efros. "Context as supervisory signal: discovering objects with predictable context." In ECCV, pp. 362–377, (2014).

[13] Weimer, Daniel, A. Y. Bengollo, and Michael Freitag. "Context-aware Deep Convolutional Neural Networks for Industrial Inspection." *Intel*. (2015).